\documentclass[runningheads]{llncs}

\usepackage[T1]{fontenc}

\usepackage{graphicx} 
\usepackage{tikz}
\usepackage{pgf}
\usepackage{amsmath, amssymb, mathtools}
\usepackage{subcaption}
\usepackage{caption} 
\DeclarePairedDelimiterX{\DKL}[2]{D_{\text{KL}}[}{]}{#1\;\delimsize\|\;#2}
\DeclarePairedDelimiterX{\E}[1]{\mathbb{E}[}{]}{#1}
\DeclarePairedDelimiterX{\E_sub}[2]{\mathbb{E}_{#1}[}{]}{#2}
\usepackage{color}
\usepackage{todonotes}
\usetikzlibrary{arrows, positioning, backgrounds,automata,snakes,shapes,petri}
 
\begin{document}

\title{Belief sharing: a blessing or a curse}
%
%
\author{Ozan \c{C}atal\inst{1} 
    \and Toon Van de Maele\inst{1} 
    \and Riddhi J. Pitliya\inst{1,2}
    \and Mahault Albarracin\inst{1,3}
    \and Candice Pattisapu\inst{1}
    \and Tim Verbelen\inst{1}}
\authorrunning{O. \c{C}atal et al.}
\institute{VERSES Research Lab, Los Angeles, California, 90016, USA \and
Department of Experimental Psychology, University of Oxford, Oxford, UK \and
Department of Computer Science, Universit\'{e} du Qu\'{e}bec \`{a} Montr\'{e}al, Montr\'{e}al, Canada \\
\email{ozan.catal@verses.ai}}
\maketitle              

\begin{abstract}
When collaborating with multiple parties, communicating relevant information is of utmost importance to efficiently completing the tasks at hand. Under active inference, communication can be cast as sharing beliefs between free-energy minimizing agents, where one agent's beliefs get transformed into an observation modality for the other. However, the best approach for transforming beliefs into observations remains an open question. In this paper, we demonstrate that naively sharing posterior beliefs can give rise to the negative social dynamics of echo chambers and self-doubt. We propose an alternate belief sharing strategy which mitigates these issues.
\end{abstract}

\section{Introduction}
Communication and the emergence of language have been a cornerstone in the development of human intelligence, as they enable human collaboration at multiple communal scales~\cite{Henrich2015}. This collaborative capability hinges significantly on how agents share and process information, particularly requiring their internal beliefs about the world to be aligned~\cite{Bahrami2010,Constant2019}.
Active inference provides a compelling framework for understanding and designing such collaborative interactions in the interest of building ecosystems of intelligence~\cite{Friston2024,FristonFirth2015,Tison2021}. In this paradigm, agents minimize variational free energy~\cite{ActInfBook}, as each agent maintains a generative model of the world which it uses to make inferences about hidden states and to plan actions. Then, communication between agents at the lowest level can be conceptualized as sharing these internal beliefs, transforming the beliefs of one agent into observable data for another~\cite{Friston2023}, these beliefs can be shared directly as we will demonstrate in this paper, but could also be present in the environment more permanently in the form of scripts\cite{Albarracin2021} and texts~\cite{Gallagher2018,Bouizegerane2020}. This belief-sharing mechanism is intended to facilitate a more coherent and efficient joint exploration of the environment.

However, translating and sharing beliefs has its challenges, as the messages shared are typically colored by one's personal priors and biases~\cite{Albarracin2022}. We found that naively sharing posterior beliefs can inadvertently lead to detrimental social dynamics, such as echo chambers, in which agents reinforce each other's biases, and self-doubt, in which agents discount their observations to favor shared, yet incorrect, beliefs. These phenomena can significantly impair the collective performance of the agents, highlighting the need for more sophisticated strategies in belief communication.
In this paper, we explore these dynamics in depth. We begin by modeling the communication between agents as belief sharing under the active inference framework, demonstrating the pitfalls of straightforward belief sharing. We then propose an alternative strategy that mitigates these issues by adjusting how beliefs are communicated. Specifically, we advocate for sharing likelihood information rather than posterior beliefs, treating other agents' observations as additional independent sources of information. This approach aims to harness the benefits of collaborative inference while avoiding the pitfalls of misleading belief reinforcement.

Our contributions are threefold:
(i) We provide a detailed analysis of how naive belief-sharing can lead to echo chambers and self-doubt.
(ii) We propose a novel communication strategy that mitigates these issues by sharing likelihoods.
(iii) We validate our approach through simulations, demonstrating improved performance and robustness in collaborative tasks.
The following sections outline our active inference model for communication, describe the experimental setup used to test our hypotheses, present our findings on echo chambers and self-doubt, and discuss our proposed solution and its implications for designing collaborative AI systems.


\section{An active inference model for communication}
In this section, we provide a summary overview of active inference, and how it can be adopted to model communication between agents. For a more in depth overview of active inference we refer the reader to~\cite{ActInfBook}.

\subsection{Perception and planning as inference}
Active inference posits that agents entertain a generative model of the environment they operate in, and casts perception and action as Bayesian inference~\cite{ActInfBook}. In general, the agent's generative model can be written as the joint probability distribution over states $s$, observations $o$ and actions $a$, with tilde denoting a time sequence of those over timesteps $t$:

\begin{equation} 
P(\tilde{s}, \tilde{o}, \tilde{a}) = P(s_0) \prod_{t}
P(o_t| s_{t})P(s_t|s_{t-1}, a_{t-1}) P(a_{t-1})  
\label{eq1}
\end{equation}

Perception now becomes inferring the posterior distributions of states given the performed actions and observations. As this is typically intractable, agents resort to variational Bayesian inference, where an approximate posterior $Q(\tilde{s}|\tilde{o})$ is optimized instead, by minimizing the variational Free Energy:

\begin{equation}
\begin{aligned}
F & = \underbrace{D_{KL}[Q(\tilde{s}|\tilde{o}))||P(\tilde{s}|\tilde{a},\tilde{o})]}_\text{posterior approximation} -\underbrace{\log P(\tilde{o})}_\text{log evidence}\\
 & = \underbrace{D_{KL}[Q(\tilde{s}|\tilde{o}))||P(\tilde{s},\tilde{a})]}_\text{complexity} - \underbrace{\mathbb{E}_{Q(\tilde{s}|\tilde{o})}[\log P(\tilde{o}| \tilde{s})]}_\text{accuracy}
\end{aligned}
\label{eq:F}
\end{equation}

It is clear that minimizing variational Free Energy is equivalent with maximizing a bound on the (log) evidence or ELBO~\cite{Bishop2007}, and encourages the model to maximize accuracy with minimal complexity.

To interact with the environment, an agent also needs to select a sequence of actions or policy $\pi = \{a_{t}, a_{t+1}, ...\}$ to execute. In active inference, planning is also treated as inference, assuming that agents will prefer policies that minimize expected Free Energy $G$. More specifically, policies are selected from

\begin{equation}
\begin{aligned}
P(\pi) & = \sigma(- G(\pi)) \text{, with} \\
G(\pi) &= \sum_{\tau=t+1}^{T} \underbrace{\mathbb{E}_{Q(o_\tau|\pi)}\big[D_{KL}[Q(s_\tau|o_\tau,\pi)||Q(s_\tau|\pi)]\big]}_{\text{(negative) Information Gain}} - \underbrace{\mathbb{E}_{Q(o_\tau|\pi)}\big[\log P(o_\tau) \big]}_{\text{Utility}}
\end{aligned}
\label{eq:G}
\end{equation}

Here, $\sigma$ denotes the softmax function, and the expected Free Energy balances information gain with some prior preference distribution over future outcomes or utility.

\begin{figure}
    \centering
    \includegraphics[width=10cm]{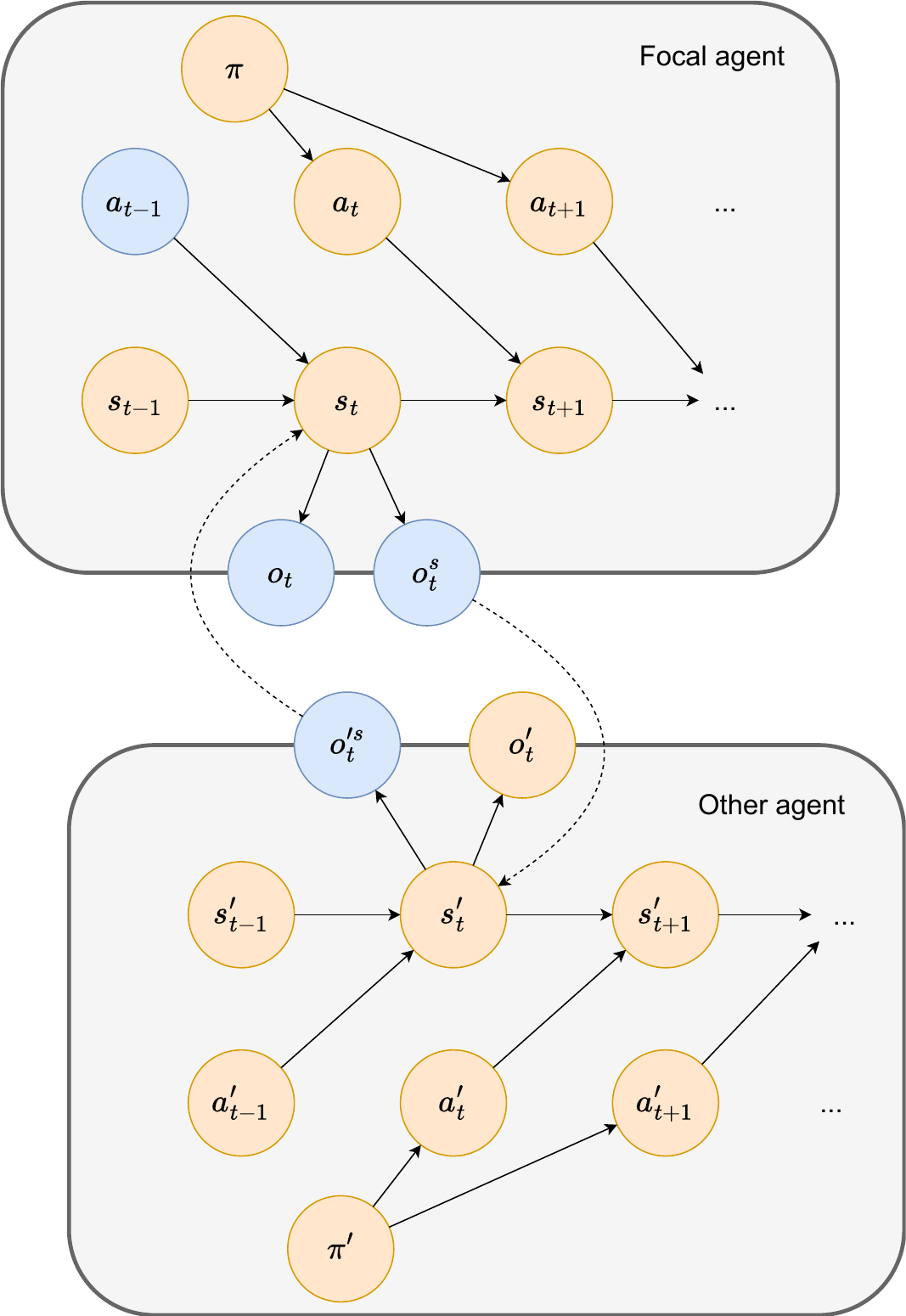}
    \caption{\textbf{Two active inference agents sharing beliefs.} The generative model of each agent is a POMDP, where observations are generated from a hidden state $s_t$. Actions $a_t$, generated from a policy $\pi$, transition this state. In addition to the typical observation $o_t$ at each timestep $t$, each agent also receives a shared observation $o^{s}_t$ that is generated from the other agent's internal beliefs $s'_t$. Blue variables are observed from the perspective of the focal agent, i.e. they observe their own actions, observations and the observations shared with the other agent.}
    \label{fig:belief-share}
\end{figure}

\subsection{Communication as belief sharing}

When agents share a common world and world model, they can benefit from sharing beliefs among each other~\cite{Friston2023}. The most straightforward way to realize this would be to share the agent's respective posterior beliefs on some shared modality. In order to achieve this, the agents generative model is expanded as shown in Fig.~\ref{fig:belief-share}. In particular, any agent, for example., the primary focal agent, assumes other agents with a similar generative model will communicate information about its beliefs of the world. To do so, we equip the focal agent with an extra observation modality $o^s_t$. Instead of being observed from the environment, $o^s_t$ is an observation generated by another agent based on its internal beliefs $s'_t$. This approach is the kind of model posited in earlier work~\cite{Friston2023}.

To realize posterior belief-sharing, agents require a likelihood mapping between posterior beliefs about latent states that are shareable among agents, i.e., $s_t$, and this observation modality $o^s_t$. In natural systems, these can be a very complex likelihood mapping, e.g., language~\cite{Friston2020} or birdsong~\cite{Friston2015}. However, in the case of AI agents, we can decide on the communication channel ourselves. One particular naive choice is to directly share the sufficient statistics of one's internal beliefs $s'_t$ and integrate these with an identity likelihood mapping, as used in~\cite{Friston2023}. However, in the remainder of this paper, we will demonstrate the fallacies of using this approach and propose a different format for shared messages.

\section{Experimental setup}


\begin{figure}[t!]
    \centering
    \begin{subfigure}[T]{0.39\textwidth}
        \centering
        \includegraphics[width=\linewidth]{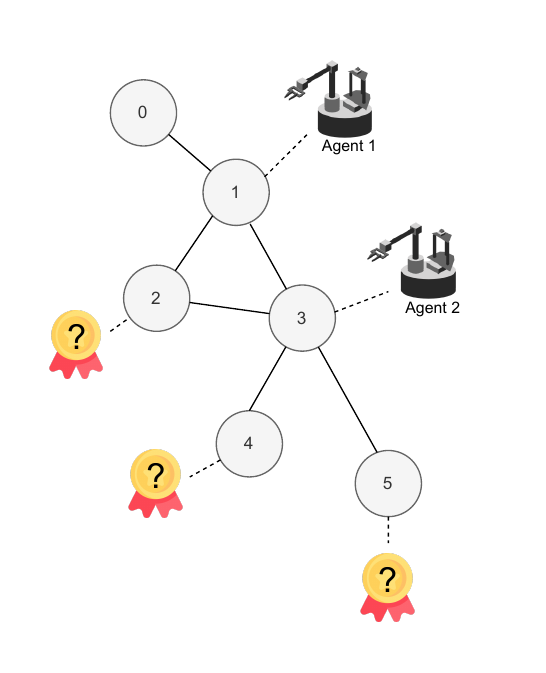}
        \caption{
        }
        \label{fig:problem-description}
    \end{subfigure}
    \begin{subfigure}[T]{0.59\textwidth}
        \centering
        \includegraphics[width=\linewidth]{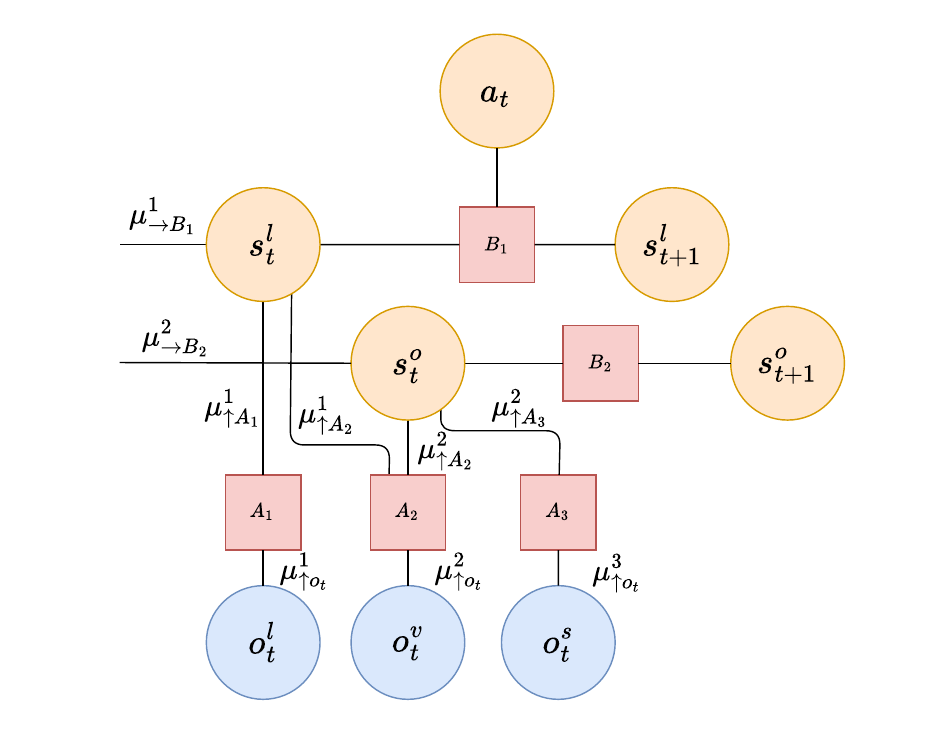}
        \caption{
        }
        \label{fig:model-detail}
    \end{subfigure}
    \caption{\textbf{Illustration of the graph environment and the agent's factor graph}. (a) agents are located on a connected graph of locations and need to find a rewarding object that might be present at one of the locations. (b) a factor graph representation of the agent's generative model. Two latent state factors that model the agent's location and the object's location respectively, give rise to two sensory modalities through a likelihood factor: the agent's location ($A_1$) and whether the object is visible ($A_2$). In addition, agents can share beliefs about the object location through belief sharing ($A_3$). The agent's location can change conditioned on move actions ($B_1$), whereas the object is kept static in our experiments ($B_2 = I$).}
    \label{fig:belief-share-factor}
\end{figure}

To demonstrate multi-agent belief sharing, we simulate an object-finding task, where multiple agents search for a rewarding object in the same environment, and can potentially share beliefs on where they think the object is. The setup and generative model for this task is depicted in Figure~\ref{fig:belief-share-factor}.
The world is represented by a graph of $N$ locations that can be visited by the agents, and agents can move between connected nodes in the graph. Each agent has two state factors, a $\text{Categorical}(N)$ variable $s_t^{l}$ which is the belief about the agent's location, and a $\text{Categorical}(N)$ variable $s_t^{o}$ which is the belief about the object location. An agent can perform one of $N$ move actions $a$, modeled using the three-dimensional dynamics tensor $B_1$. 

\begin{equation*}
    B_1^{i,j,a} = \begin{cases}
        1.0,& \text{if } a = i \wedge \text{connected}(i,j)\\
        1.0,& \text{if } i=j \\
        0.0,& \text{otherwise}
    \end{cases}    
\end{equation*}

We assume the object is static, i.e. its dynamics $B_2$ are modeled by the identity matrix $I$. The agent has three observation modalities. First, it observes its location $o_t^{l}$, with a near identity likelihood mapping to the location state factor:

\begin{equation*}
    A_1^{i,j} = \begin{cases}
        0.99,& \text{if } i = j\\
        0.01,& \text{otherwise }
    \end{cases}
\end{equation*}
Second, it observes $o_{t}^{v}$ whether the object is visible or not. This is governed by a three-dimensional likelihood mapping $A_2$ where

\begin{equation*}
    A_2^{v, i, j} = \begin{cases}
        0.2,& \text{if } i = j \wedge v = \text{not visible}\\
        0.8,& \text{if } i \neq j \wedge v = \text{not visible}\\
        0.8,& \text{if } i = j \wedge v = \text{visible} \\
        0.2,& \text{if } i \neq j \wedge v = \text{visible}
    \end{cases}
\end{equation*}
Finally, there is the belief-sharing observation $o_t^{s}$ which contains the shared information from the other agent.

The agents are initialized with a prior on $s_t^{l}$ set to their starting location, and a prior on $s_t^{o}$ set to the initial belief on where to find the object. This is typically set uniform to (a subset of) the available locations to foster searching behavior. The preference $C$ is to have the object visible outcome.

\section{Echo chambers}

\begin{figure}[b!]
    \centering
    \includegraphics[width=.35\linewidth]{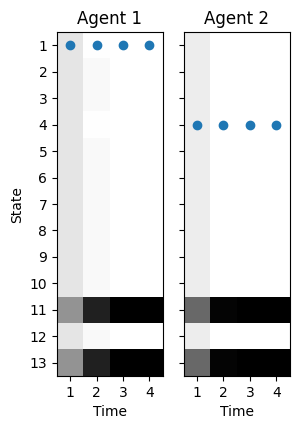}
    \caption{\textbf{Simulation of an echo-chamber}. We initialize both agents with a small prior belief that the object will be present at location 11 or 13. Then, we let the agents share their beliefs. Note that this reinforces the belief that the object will be at either one of the locations. The next columns show the evolution of both agents where they keep observing the environment. We see that in the transition from time 1 to time 2 the agents increase the belief that the object is at the a priori believed location even though there is no new evidence to support this belief. Agent location is depicted using the blue dots in both panels.}
    \label{fig:echo-chamber}
\end{figure}

In a model which shares the posterior beliefs of one agent as observations of another, Bayesian model updating reinforces redundant priors shared among them. The consequent simulated behavior mirrors the “echo chamber effect" wherein messages communicated by like-minded agents are amplified and returned. Psychological interpretations of the echo chamber effect are illustrated in the consequences of social media feed algorithms, which are often engineered to encourage user engagement with sympathetic posts~\cite{Cinelli2021}. An algorithmically curated social media curriculum increases engagement by anticipating a user's expected social media observations and fulfilling those expectations. Promoted content thereby constructs homophilic interaction networks which facilitate the construction and reinforcement of shared narratives among users. In worst case scenarios, the result is the unimpeded flow of misinformation on social media platforms. More generally, shared narratives facilitated by social media feed algorithms result in an increase in confidence of posterior beliefs even when external evidence is absent or intentionally excluded. 

Ignoring new evidence can be adaptive, such as when this strategy facilitates in-group cooperation~\cite{Kim2020}.
However, negative consequences also result, such as when outside sources are discredited or distrusted. 

Corresponding to the echo chamber effect, one pitfall of belief sharing is that when agents have established even small prior beliefs on their goal, if those priors are shared, then an echo chamber forms. The agents, however, reinforce their prior beliefs through the communication method, resulting in ever-increasing beliefs on the goal location even when there is no new evidence to support this belief. Fig.~\ref{fig:echo-chamber} shows a simulation triggering this situation. The agents start with a small belief that the object will be present at two locations within the graph to simulate a longer-running experiment where such a situation might naturally occur. We have restricted the movement of the agents and prohibited the agents from accumulating more evidence about the object's actual location. However, the agents keep increasing their belief on the object being present at the a priori believed location because of the constant sharing of beliefs. Once the agents create such an echo chamber, more often than not, they are stuck in this faulty belief unless they all sufficiently change their belief at the same time.

\section{Self-doubt}

\begin{figure}[t!]
    \centering
    \includegraphics[width=.7\linewidth]{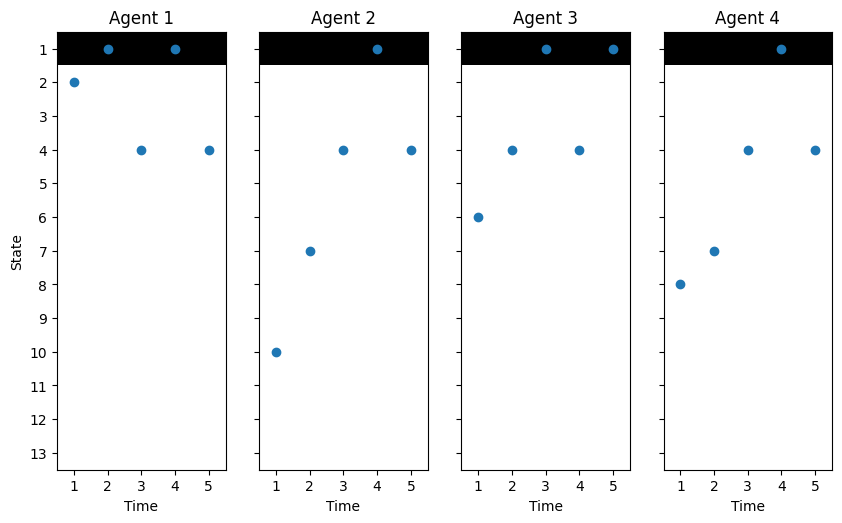}
    \caption{\textbf{Simulation of self-doubt for 4 agents.} 
    Again, each panel displays the evolution of the posterior belief as a function of time, where darker colors indicates a higher degree of belief. Blue dots indicate the agent location. All agents are initialized with a strong prior belief that the object will be at location 1, as indicated by the dark shaded area. This reflects a potential situation where all agents have been acting in the environment for a long time, accruing faulty evidence. In this case the communication mechanism prohibits the agents from discovering that the object is not there, even after observing its absence multiple times.}
    \label{fig:self-doubtl}
\end{figure}

In an echo chamber, a belief sharing agent's confidence about their priors is reinforced in the absence of external evidence. Conversely, self-doubt refers to a scenario in which the agent's self-confidence is degraded as a result of belief sharing. In our simulations, the paradigmatic example is when multiple agents are fixated on a search task with complimentary plans for exploring the environment. Naturally, siblings faulty beliefs and ignore all evidence that points to the contrary. This can occur after an echo chamber is formed but could also occur independently. In this scenario, the agents again reinforce each other's beliefs in such a strong way that the agents ``doubt'' their observations originating from the environment. Fig.~\ref{fig:self-doubtl} visually overviews a simulation showcasing this phenomenon. The agents are initialized with a strong belief on the object location. In this particular simulation, the agents have a single peaked belief on the object location (location 1 in the graph), which could occur after an echo chamber situation. The agents are unrestricted in their movement as long as they follow the underlying graph structure. We see that even though eventually all agents visit location 1, they cannot correctly eliminate that location as a possible object location. The incoming beliefs of the other agents overrule their sensory observations.


\section{To share or not to share?}

Given that sharing agents' beliefs can give rise to the aforementioned issues, it is worth wondering what information can best be shared to allow multi-agent cooperation in active inference agents. In particular, the update rule for $s^o_t$, written in variational message passing notation~\cite{Winn2005}, is of the form
\begin{equation}
\begin{aligned}
    s^o_t &= \sigma( \mu^2_{\rightarrow B_2} + \mu^2_{\uparrow A_2} + \mu^2_{\uparrow A_3}),  \\
    \mu^f_{\uparrow A_g} &= o^g_t \odot \varphi(\textbf{a}^g) \odot_{i \in pa(g) \backslash f} s^i_t , 
\end{aligned}
\label{eq:VMP}
\end{equation}

where $\sigma$ is the softmax function, $\mu^f_{\uparrow A_g}$ the message from observation modality $g$ to state factor $f$ and $\varphi(\textbf{a}^g)$ the digamma function of the Dirichlet counts corresponding to the parameters of the likelihood model~\cite{Friston2023}. Effectively, the update message is comprised of a part coming from a prior given by our beliefs on the previous timestep $\mu^2_{\rightarrow B_2}$, a part based on the latest observation $\mu^2_{\uparrow A_2}$, and a part communicated by the other $\mu^2_{\uparrow A_3}$. When we communicate the other's posterior parameters directly through an identity likelihood mapping, we have $\mu^2_{\uparrow A_3} = \mu^{2,other}_{\rightarrow B_2} + \mu^{2,other}_{\uparrow A_2}$. This shows that, indeed, when agents have similar priors, this gets double counted in the belief update.

To address this, we will now instead share the other's likelihood message only, i.e. $\mu^2_{\uparrow A_3} = \mu^{2,other}_{\uparrow A_2}$. This scheme leaves out agents' prior beliefs about the state and only shares the agent's interpretation of the observation, treating the other agents as extra independent observers for the exact latent cause in the world. We call this scheme `likelihood sharing`.

\begin{figure}[h]
    \centering
    \includegraphics[width=.35\linewidth]{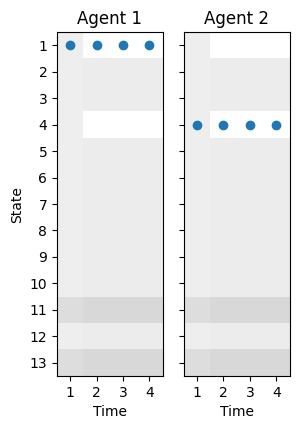}
    \caption{\textbf{Illustration of the lack of echo-chamber like behaviour when sharing likelihoods.} In this figure, the situation leading to an echo chamber is recreated. Both agents are initialized with the same prior belief that the object is most likely at locations 11 and 13; however, because of the sharing of likelihoods, they do not get stuck in an echo chamber and do not increase the beliefs when there is no new evidence.}
    \label{fig:echo-chamber-fix}
\end{figure}

In the particular scenario of object-finding agents in a graph world, the agents share their current location and visibility observation as passed through their object location A-tensor; each receiving agent can integrate this observation quite easily into their own posterior belief by using Bayes rule. In effect, each agent treats the other agents as an extra ``pair of eyes'' in the search for the object, inferring their posterior update if they observed what the other agent had observed.

\begin{figure}[h]
    \centering
    \includegraphics[width=.7\linewidth]{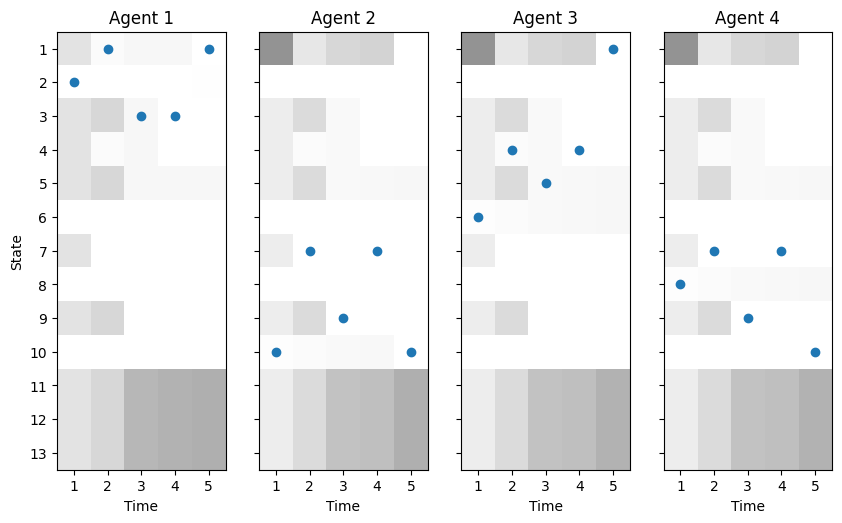}
    \caption{\textbf{Alleviation of the self-doubt behavior under likelihood-sharing.} As with the previous depiction of this scenario, the agents are initialized with a strong belief that the object will be at location 1; nonetheless, due to the different communicated belief, the agent no longer ignores the evidence for the object's absence.}
    \label{fig:self-doubt-fix}
\end{figure}

In Fig.~\ref{fig:echo-chamber-fix} and Fig \ref{fig:self-doubt-fix}, the same simulations from earlier are reprised but using the new likelihood-sharing mechanism. The echo chamber and self-doubt phenomena are no longer present even when providing the same initial conditions.

\begin{figure}
    \centering
    \includegraphics[width=0.4\textwidth]{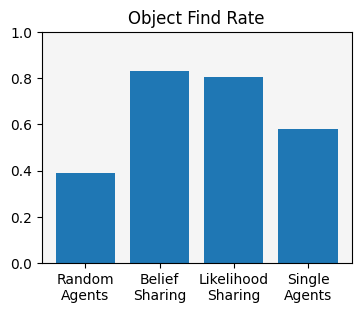}
    \caption{\textbf{Overview of the average object find rate for each type of agent.} We measured the percentage of finding the object by any of the two agents over all possible start configuration in the environment. This is repeated five times to account for variability in the action selection process.}
    \label{fig:results}
\end{figure}

Finally, Figure~\ref{fig:results} compares likelihood sharing to belief sharing and not communicating at all. All methods are tested over all possible combinations of agent starting locations and object locations in the environment, and each configuration is repeated five times. The trials are evaluated on the percentage of times the object was found. From the experiments, the likelihood-sharing agents are on par with the naive belief-sharing agents. At the same time, they both outperform the random agents and the non-communicating agents. 

\section{Discussion}
The results presented in this paper highlight the potential pitfalls of belief-sharing in multi-agent systems under the framework of active inference. Our findings suggest that naive sharing of posterior beliefs can lead to undesirable social dynamics such as echo chambers and self-doubt, severely hampering the agents' performance in collaborative tasks.

Echo chambers in human societies are well-documented phenomena where groups of individuals reinforce their preconceptions, often without external validating evidence. In our simulations, a similar effect occurs when agents continuously share their posterior beliefs. Initial biases can get amplified through repetitive belief sharing, leading to overly confident but potentially erroneous shared beliefs. This results in the agents becoming overconfident in incorrect hypotheses, thereby hampering their search or exploration processes.
Similarly, self-doubt arises when agents' observations contradict the reinforced shared beliefs, leading them to disregard their sensory inputs. This mirrors real-world psychological effects where individuals question their perceptions in the face of strong peer influence. In our simulations, agents maintained strong incorrect beliefs about the object's location despite direct evidence to the contrary due to the influence of shared but incorrect posterior beliefs.

Our proposed strategy of sharing likelihoods rather than posterior beliefs mitigates the issues of echo chambers and self-doubt. By sharing interpreted observations rather than fully formed beliefs, agents can integrate new information without being overwhelmed by the potentially erroneous priors of others. This approach allows agents to utilize each other as additional sensing mechanisms, providing independent evidence that can be more robustly combined with their observations.
By treating other agents' observations as additional data points rather than beliefs, the system remains more flexible and resilient to individual errors. The proposed approach was only validated in simulated experiments but might also apply to more general and complex scenarios with further modifications.

Our work suggests that the type of information shared among active inference agents must be carefully considered to avoid counterproductive dynamics. Future research can build on these insights by exploring other belief-sharing strategies and their impacts on system performance. Additionally, exploring these dynamics in more complex and varied environments, including those with adversarial elements, could provide deeper insights into the performance of different communication strategies. As our approach assumed full and honest collaboration between the agents, another future avenue of research would be to investigate the impact of dishonesty and, consequently, the discounting of communications between untrusted agents.

In conclusion, while belief sharing among active inference agents can enhance collaborative performance, it also risks reinforcing incorrect beliefs and undermining individual observations. Our proposed likelihood-sharing mechanism offers a promising solution by leveraging the strengths of collective sensing while mitigating the pitfalls of echo chambers and self-doubt without significant changes to the underlying model. Such strategies will be essential for developing robust, efficient, and adaptive collaborative agents as multi-agent active inference systems are designed.


\bibliographystyle{ieeetr}
\bibliography{references}

\begin{thebibliography}{10}

\bibitem{Henrich2015}
J.~Henrich, {\em The Secret of Our Success: How Culture Is Driving Human Evolution, Domesticating Our Species, and Making Us Smarter}.
\newblock Princeton University Press, Oct. 2015.

\bibitem{Bahrami2010}
B.~Bahrami, K.~Olsen, P.~E. Latham, A.~Roepstorff, G.~Rees, and C.~D. Frith, ``Optimally interacting minds,'' {\em Science}, vol.~329, p.~1081–1085, Aug. 2010.

\bibitem{Constant2019}
A.~Constant, M.~J.~D. Ramstead, S.~P.~L. Veissière, and K.~Friston, ``Regimes of expectations: An active inference model of social conformity and human decision making,'' {\em Frontiers in Psychology}, vol.~10, Mar. 2019.

\bibitem{Friston2024}
K.~J. Friston, M.~J. Ramstead, A.~B. Kiefer, A.~Tschantz, C.~L. Buckley, M.~Albarracin, R.~J. Pitliya, C.~Heins, B.~Klein, B.~Millidge, D.~A. Sakthivadivel, T.~St~Clere~Smithe, M.~Koudahl, S.~E. Tremblay, C.~Petersen, K.~Fung, J.~G. Fox, S.~Swanson, D.~Mapes, and G.~René, ``Designing ecosystems of intelligence from first principles,'' {\em Collective Intelligence}, vol.~3, Jan. 2024.

\bibitem{FristonFirth2015}
K.~J. Firston and C.~D. Firth, ``Active inference, communications and hermeneutics,'' {\em Cortex}, vol.~68, pp.~129--143, 2015.

\bibitem{Tison2021}
R.~Tison and P.~Poirier, ``Active inference and cooperative communication: An ecological alternative to the alignment view,'' {\em Frontiers in Psychology}, vol.~12, 2021.

\bibitem{ActInfBook}
T.~Parr, G.~Pezzulo, and K.~J. Friston, {\em {Active Inference: The Free Energy Principle in Mind, Brain, and Behavior}}.
\newblock The MIT Press, 03 2022.

\bibitem{Friston2023}
K.~J. Friston, T.~Parr, C.~Heins, A.~Constant, D.~Friedman, T.~Isomura, C.~Fields, T.~Verbelen, M.~Ramstead, J.~Clippinger, and C.~D. Frith, ``Federated inference and belief sharing,'' {\em Neurosci. Biobehav. Rev.}, p.~105500, Dec. 2023.

\bibitem{Albarracin2021}
M.~Albarracin, A.~Constant, K.~J. Friston, and M.~J.~D. Ramstead, ``A variational approach to scripts,'' {\em Frontiers in Psychology}, 2021.

\bibitem{Gallagher2018}
S.~Gallagher and M.~Allen, ``Active inference, enactivism and the hermeneutics of social cognition,'' {\em Synthese}, vol.~195, no.~6, pp.~2627--2648, 2018.

\bibitem{Bouizegerane2020}
N.~Bouizegarene, M.~Ramstead, A.~Constant, K.~Friston, and L.~Kirmayer, ``Narrative as active inference,'' {\em Frontiers in Psychology}, vol.~15, 2024.

\bibitem{Albarracin2022}
M.~Albarracin, D.~Demekas, M.~J.~D. Ramstead, and C.~Heins, ``Epistemic communities under active inference,'' {\em Entropy}, vol.~24, p.~476, mar 2022.

\bibitem{Bishop2007}
C.~M. Bishop, {\em Pattern Recognition and Machine Learning (Information Science and Statistics)}.
\newblock Springer, 1~ed., 2007.

\bibitem{Friston2020}
K.~J. Friston, T.~Parr, Y.~Yufik, N.~Sajid, C.~J. Price, and E.~Holmes, ``Generative models, linguistic communication and active inference,'' {\em Neuroscience \&; Biobehavioral Reviews}, vol.~118, p.~42–64, Nov. 2020.

\bibitem{Friston2015}
K.~J. Friston and C.~D. Frith, ``Active inference, communication and hermeneutics,'' {\em Cortex}, vol.~68, p.~129–143, July 2015.

\bibitem{Cinelli2021}
M.~Cinelli, G.~De~Francisci~Morales, A.~Galeazzi, W.~Quattrociocchi, and M.~Starnini, ``The echo chamber effect on social media,'' {\em Proceedings of the National Academy of Sciences}, vol.~118, Feb. 2021.

\bibitem{Kim2020}
M.~Kim, B.~Park, and L.~Young, ``The psychology of motivated versus rational impression updating,'' {\em Trends in Cognitive Sciences}, vol.~24, p.~101–111, Feb. 2020.

\bibitem{Winn2005}
J.~Winn and C.~M. Bishop, ``Variational message passing,'' {\em Journal of Machine Learning Research}, vol.~6, no.~23, pp.~661--694, 2005.

\end{thebibliography}

\end{document}